\documentclass[letterpaper, 10 pt, conference]{ieeeconf}  

\IEEEoverridecommandlockouts                              
\usepackage{multirow}
\usepackage{multicol}
\usepackage{booktabs}
\usepackage[table]{xcolor}
\usepackage{float}

\usepackage{graphicx}
\usepackage{duckuments}
\usepackage{amsmath}

\usepackage{xcolor}
\usepackage{todonotes}

\usepackage{pifont}
\usepackage{bbding}

\usepackage{siunitx}

\usepackage[acronym]{glossaries}
\glsdisablehyper
\newacronym{vlm}{VLM}{Vision-Language Model}
\newacronym{icl}{ICL}{In-Context Learning}
\newacronym{llm}{LLM}{Large Language Model}
\newacronym{fpv}{FPV}{First-Person View}
\newacronym{tpv}{TPV}{Third-Person View}
\newacronym{rag}{RAG}{Retrieval-Augmented Generation}
\newacronym{vqa}{VQA}{Visual Question-Answering}
\newacronym{vit}{ViT}{Vision Transformer}
\newacronym{rl}{RL}{Reinforcement Learning}
\newacronym{mmr}{MMR}{Maximal Marginal Relevance}

\usepackage{subcaption}

\usepackage{lipsum}

\usepackage{crossreftools}

\makeatletter
\let\NAT@parse\undefined
\makeatother

\usepackage[hidelinks]{hyperref}
\newcommand\rurl[1]{%
\texttt{\href{http://#1}{\nolinkurl{#1}}}%
}
\usepackage[capitalise]{cleveref}

\title{\LARGE \bf
Select2Plan: Training-Free ICL-Based Planning \\through VQA and Memory Retrieval
}
\author{Davide Buoso$^{*1 2}$, Luke Robinson$^1$, Giuseppe Averta$^2 $, Philip Torr$^1$, Tim Franzmeyer$^{1\dagger}$, Daniele De Martini$^{1\dagger}$
\\
$^{1}$Mobile Robotics Group (MRG) and Torr Vision Group (TVG), University of Oxford, UK.\\
$^{2}$Politecnico di Torino, Italy.\\
\thanks{$^{*}$ Corresponding author. Please contact at the email address
\texttt{davide.buoso@polito.it} \newline $^\dagger$ Equal supervisory contribution.}
\thanks{
This work was supported by EPSRC Impact Acceleration Account (IAA) ``Robotics  Inversion''.}
\thanks{ This study was carried out also within the Future Artificial Intelligence Research (FAIR) and received funding from the European Union Next-GenerationEU (PIANO NAZIONALE DI RIPRESA E RESILIENZA (PNRR) – MISSIONE 4 COMPONENTE 2, INVESTIMENTO 1.3 – D.D. 1555 11/10/2022, PE00000013). 
}%
}

\renewcommand{\baselinestretch}{0.977}
\begin{document}

\maketitle

\begin{abstract}
This study explores the potential of off-the-shelf \glspl{vlm} for high-level robot planning in the context of autonomous navigation.
Indeed, while most of existing learning-based approaches for path planning require extensive task-specific training/fine-tuning, we demonstrate how such training can be avoided for most practical cases. 
To do this, we introduce Select2Plan (S2P), a novel training-free framework for high-level robot planning which completely eliminates the need for fine-tuning or specialised training. 
By leveraging structured \gls{vqa} and \gls{icl}, our approach drastically reduces the need for data collection, requiring a fraction of the task-specific data typically used by trained models, or even relying only on online data.
Our method facilitates the effective use of a generally trained \gls{vlm} in a flexible and cost-efficient way, and does not require additional sensing except for a simple monocular camera.
We demonstrate its adaptability across various scene types, context sources, and sensing setups.
We evaluate our approach in two distinct scenarios: traditional \gls{fpv} and infrastructure-driven \gls{tpv} navigation, demonstrating the flexibility and simplicity of our method.
Our technique significantly enhances the navigational capabilities of a baseline \gls{vlm} of approximately 50\% in \gls{tpv} scenario, and is comparable to trained models in the \gls{fpv} one, with as few as 20 demonstrations.

\end{abstract}

\glsresetall

\section{Introduction}

Path planning for vehicles is a longstanding problem in robotics, traditionally addressed using model-based or \gls{rl} approaches \cite{staroverov2020real,visualrlzhou2021,shah2023gnm}.
However, methods that directly learn from experience often struggle when confronted with ambiguous or unfamiliar scenarios.
Interestingly, recent research has shown that \glspl{llm} and \glspl{vlm}, demonstrate surprising reasoning capabilities that can be adapted for proposing robot paths in arbitrary scenes \cite{zeng2024perceivereflectplandesigning}.
Indeed, these models excel at incorporating common-sense reasoning acquired during their long pretraining phase \cite{radford2019language}. 
This ability is crucial in robotics operations, where the deployment scenario rarely aligns perfectly with the training dataset \cite{williams2024masked,williams2024mitigating}.
While methods like LoRA \cite{hu2021lora} reduce the computational cost of fine-tuning \glspl{llm} and \glspl{vlm}, they still require domain-specific data, which can be costly to obtain.
In parallel, \gls{icl} and \gls{rag} have shown promising results in scoping the ability of \glspl{llm} \textit{at deployment time} with no additional fine-tuning, mitigating these costs.
\begin{figure}[t]
    \centering
    \includegraphics[width=0.9\linewidth]{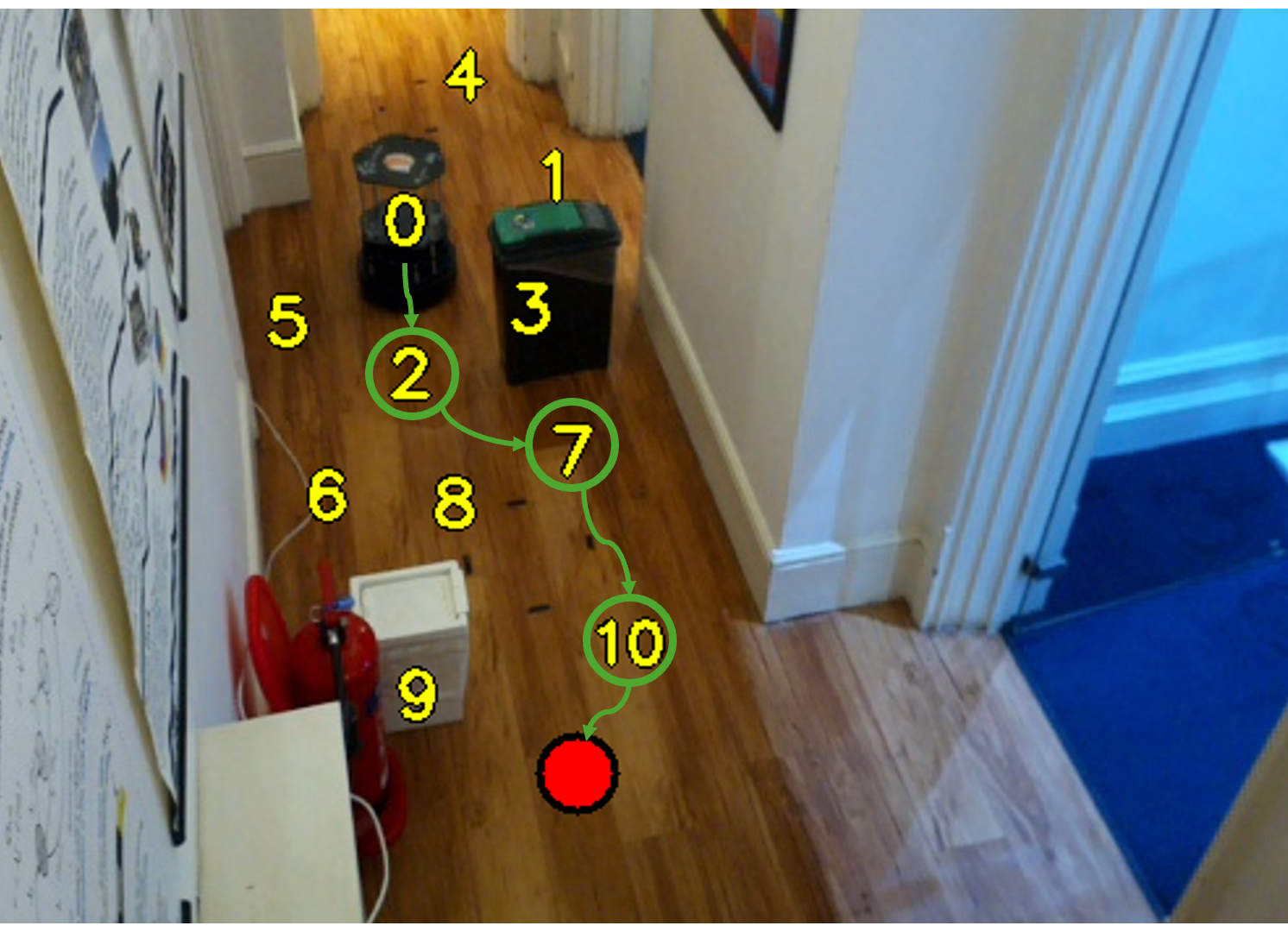}
    \caption{High-level demonstration of S2P in a \acrshort{tpv} scenario.
    The robot must reach the red mark from its location, controlled solely via the external camera, shown in the figure.
    S2P proposes candidate keypoints -- in yellow -- and draws them into the original image before requesting a feasible trajectory to an off-the-shelf \acrshort{vlm}.
    The latter will output a trajectory -- green -- as a sequence of keypoints, ideally yielding a trajectory that avoids obstacles -- e.g. 3 and 9.    \label{fig:catchy}}
    \vspace{-20pt}
\end{figure}

Our novel framework -- Select2Plan (S2P) -- combines \gls{vqa} and \gls{icl} with \glspl{vlm} in a training-free manner, showing remarkable flexibility across various scenes, contexts, and setups.

More specifically, we formulate the planning problem as a \gls{vqa} task using visual prompting.
A high-level overview of the approach can be observed in \cref{fig:catchy}.
Inspired by \cite{nasiriany2024pivot} and \cite{sathyamoorthy2024convoi}, we generate a set of position candidates in the image space and use them as part of a query mechanism to a \gls{vlm}, to extract the next robot move.
We combine this approach with \gls{icl} to enhance the model's reliability: we retrieve similar successful samples and use them, along with the current annotated image, as context to support the model's generalization.
In this way, we can generate a robust path, which can span multiple planning steps within a single response, in contrast to the iterative approach taken by \cite{nasiriany2024pivot}.

We evaluate our system in two different navigation scenarios. 
The first is a more traditional \gls{fpv}, where the robot is equipped with a monocular camera and needs to reach specific objects in the scene.
The challenge in this case is the sensor's limited view, as the goal object might get out of view while the robot navigates the environment.
As a second test-bench, we consider a robot controlled through eye-to-hand visual servoing, as in \cite{robinson2023robot,robinson2023visual} and as depicted in \cref{fig:catchy}.
Here, the camera is not physically attached to the robot, and the far viewpoint inherently limits the depth \cite{zhong2024nerfoot} and spatial resolution.
However, given the widespread use of CCTV cameras, we believe this approach offers new opportunities and, interestingly, this setup also mirrors the type of data that \glspl{vlm} are trained on -- static RGB images paired with textual descriptions -- making these models well-suited for tasks involving external camera navigation.
We show how our setup can flexibly adapt to both visual inputs and diverse sources of context, such as videos from the Internet or even human traversal of the scenario.

To summarise, our main contributions are:
\begin{enumerate}
    \item A framework for planning and navigation using only RGB data, leveraging structured \gls{vqa}, \gls{icl} and retrieval techniques to reduce the task-related data needed to just a handful of episodes.
    \item The application of this framework to two separate scenarios: traditional \gls{fpv} navigation and infrastructure-driven \gls{tpv} navigation.
    \item An extensive analysis of the impact of different sources of in-context examples on the system's overall performance.
\end{enumerate}
Our empirical analysis demonstrates that our approach enhances the navigational abilities of \glspl{vlm} without requiring further training, and lays the groundwork for more sophisticated and flexible planning in autonomous systems.
To the best of our knowledge, our approach is the first that can seamlessly adapt to multiple setups \textit{and} utilise multiple sources of in-context samples.

\begin{figure*}[ht]
    \centering
    \begin{subfigure}{\textwidth}
    \centering
        \includegraphics[width=\textwidth]{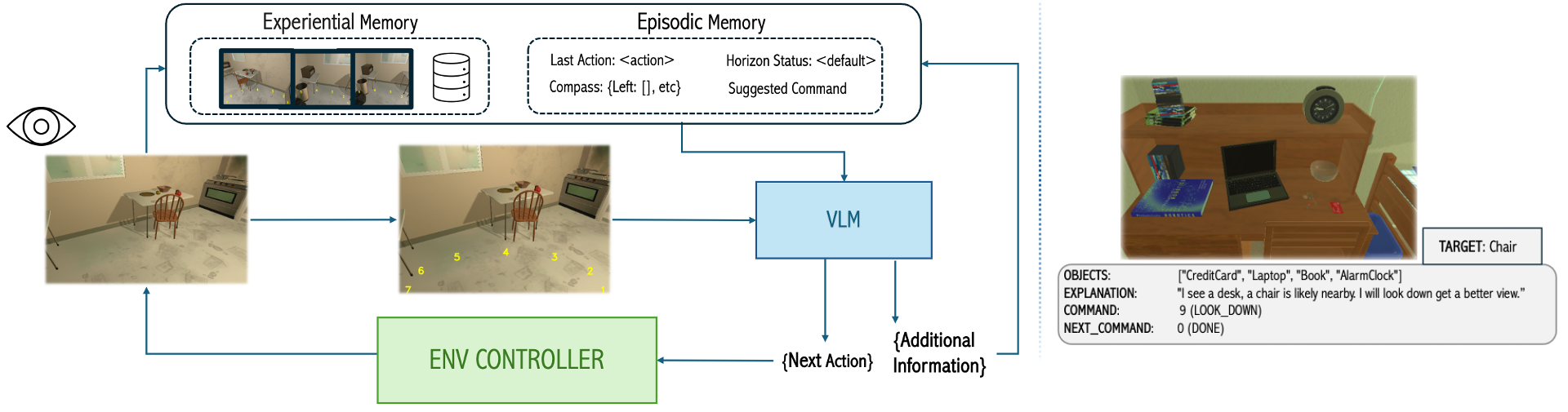}
        \caption{\label{fig:1st_frame}}
    \end{subfigure}

    \begin{subfigure}{\textwidth}
    \centering
        \includegraphics[width=\textwidth]{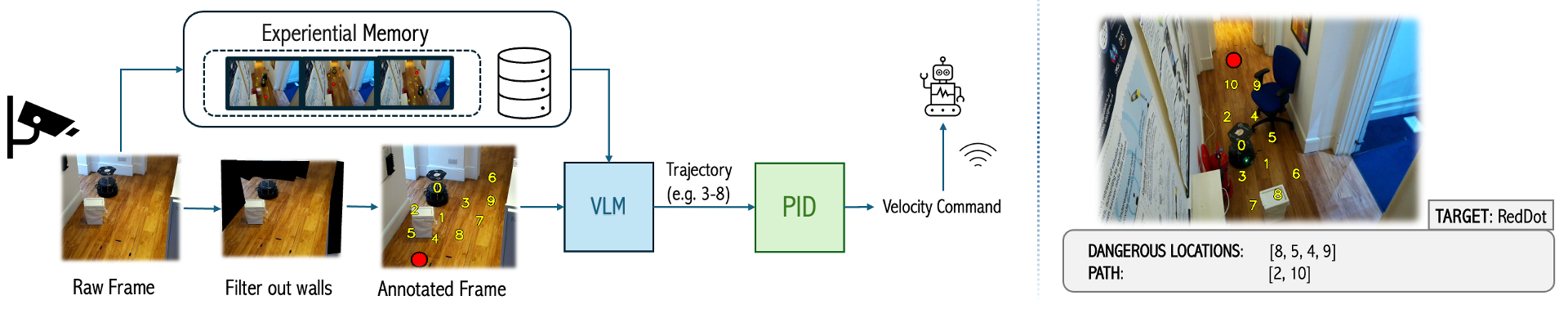}
        \caption{\label{fig:3rd_frame}}
    \end{subfigure}
    \caption{Overview of the proposed approach in \gls{fpv} (a) and \gls{tpv} (b).
    The two settings are designed to fit two specific scenarios but share their components.
    The framework takes a live image from the onboard or a CCTV camera and retrieves similar images from the experiential memory.
    It is then annotated and passed, with the sampled images and an optional episodic memory, to the \gls{vlm} to retrieve the next commands to send to the platform and explanations. 
    The main difference is the absence of an Episodic Memory in the \gls{tpv} setting, where the off-board sensing setup empirically limits its benefits. Alongside the overview, response examples are presented for both setups.
    \label{fig:methodology}
}
\vspace{-15pt}
\end{figure*}

\section{Related Work}

Our approach is at the intersection of robot navigation, planning, \glspl{llm}, \glspl{vlm}, and \gls{icl}.
Here, we review the most relevant works to highlight both similarities and differences, emphasizing the unique aspects of our methodology.

Traditional methods for robot navigation tasks, such as Object Navigation and Visual Navigation, have historically relied on \gls{rl} to train policies for complex tasks.
Works like those of \cite{staroverov2020real,visualrlzhou2021,shah2023gnm} employed deep \gls{rl} models to learn navigation policies based on visual input.
More recently, transformer-based models have emerged as a powerful alternative, often yielding better generalization due to their capacity to model long-range dependencies in data \cite{fukushima2022objectmemorytransfo,shah2023vint,sridhar2023nomad}.

Lately, the attention has shifted to the use of \glspl{llm} and \glspl{vlm} for several tasks, such as planning, navigation and manipulation \cite{zhou2024navgpt2unleashingnavigationalreasoning,duan2024manipulateanything}.
\glspl{vlm} have shown great promise for high-level decision-making in robotics, as they integrate visual perception with language-based reasoning. 
\cite{pmlr-v205-shah23b} demonstrated how a pre-trained \gls{llm} can be utilized in zero-shot settings to control robots. \cite{zeng2024perceivereflectplandesigning,duan2024manipulateanything,chen2024mapgpt} similarly explored how these models can handle real-world tasks without additional training, proving their flexibility in diverse scenarios. \gls{icl} has gained traction in tasks that require minimal data adaptation.
Recent works, such as \cite{dipalo2024kat}, have highlighted the effectiveness of \gls{icl} integrated with memory-based retrieval for robotics applications. 
However, these methods typically focus on a single setup and often require external modules for object recognition or advanced techniques to extract textual features.
Our approach diverges by adopting a zero-training pipeline, which integrates Image-Based \gls{icl} coupled with \gls{vqa} to further enhance the capabilities of \glspl{vlm}, without the need of additional techniques and collecting just an handful of samples to achieve the task.
This, allows our framework to generalize across both \gls{fpv} and \gls{tpv} setups without the need for specialized sensors or extensive pre-training.
Our model can efficiently generate navigational plans from Image-Text pairs, making it more adaptable to different scenarios and goals and completely different setups than previous methods.

\section{Methodology}
After discussing the general framework, this section details the specific declinations for the two deployment scenarios: \gls{tpv} and \gls{fpv}.
\Cref{fig:methodology} depicts the two in detail.

\subsection{Vision Language Model and In-Context Learning}

To provide context, we use a \textit{History-Injection} \gls{icl}, where a fictitious chat conversation is created.
In the chat, episodes retrieved from a memory database are split in \emph{query} (annotated image and prompt) and \emph{answer}, and injected into the model as turns of conversation as question and answer.
This yields a multi-turn conversation of $k$ turns, with $k$ equal to the number of \gls{icl} samples.
Finally, we explicitly ask the model, based on its previous responses, to process a new image.

As we can see from \cref{fig:methodology}, our proposed approach is composed of five main components: an experiential memory, a sampler, an episodic memory, an annotator and a prompt templating engine, in addition to the \gls{vlm} and the controller.
We will discuss them in the following.

\subsubsection{Experiential Memory}
The Experiential Memory is a collection of annotated images that can be used as context to influence the answer of the \gls{vlm} via \gls{icl}.
These experiences can be gathered in different ways, and we discuss different combinations in our result section, where we test with data coming from the same or different environments, or even directly processing online footage.

\subsubsection{Sampler}
The sampler aims to select the most appropriate samples from the Experiential Memory to be presented to the \gls{vlm} as context.
We feed a \gls{vit} -- a Swin Transformer \cite{liu2021swintransformerhierarchicalvision} -- with the live camera image to recover similar situations from the Experiential Memory.
We represent the image -- and the Experiential Memory samples -- as the average output of the last hidden layer and obtain a feature vector which we employ as the query.

Empirically, we observed that building a diverse context (both similar and different situations) benefits the model's generalisation to the current situation.
To balance out the similarity, we incorporated a re-ranking algorithm adapted to our framework into the retrieving process.
We employ a \gls{mmr} \cite{mmr}, which aims to reduce redundancy and increase sample diversity according to a combined criterion of query relevance and novelty of information.
\Gls{mmr} is defined as follows:
\begin{equation}
\begin{split}
\mathtt{MMR}(Q, M, C) = &\arg\max_{s_i \in M \setminus C} \bigg[ \lambda <s_i, Q> + \\
&- (1 - \lambda) \max_{s_j \in C} <s_i, s_j> \bigg]
\end{split}
\end{equation}
where $Q$ represents the query image embedding and $M$ the experiential memory.
This algorithm operates by iteratively selecting images (samples $s_i \in M$), to add to the context $C$,  based on a trade-off between two factors: the image's relevance -- similarity -- to the query and the image's dissimilarity from the samples that have already been chosen.
The goal is to ensure that each selected image adds new, informative content rather than repeating information.
The scalar $\lambda$ balances this trade-off.

\subsubsection{Episodic Memory}
We define the Episodic Memory as all the information related to the current episode that can be useful for informed navigation.
We avoid incorporating the full, raw context of the previous responses to avoid confusion, but we incorporate a text-based, simplified representation of the scene to condition the \gls{vlm}'s response.

\subsubsection{Annotator}
Before presenting the live and episodic images to the \gls{vlm}, we take inspiration from \cite{nasiriany2024pivot} and visually annotate them with the possible actions the model can choose from.
These annotations are platform-specific, and are applied as numerical values superimposed on the image frame; we will discuss them more in this section.

\subsubsection{Prompt Templating Engine}
The prompt template engine is the core of our prompt engineering step.
It combines information from the Experiential and Episodic Memories with the live, annotated image and the task at hand in a digestible form for the \gls{vlm}.
The prompt contains information about the type of data the \gls{vlm} is presented -- Experiential, Episodic and live -- how to decode the annotations into \textit{a sequence} of actions and the request to fulfil the navigation task.
We also instruct the \gls{vlm} to use a JSON format for its output to easily integrate it into any control pipeline already deployed on the robotic platform.

\subsubsection{VLM}
The \gls{vlm} is prompted to select a discrete action from the annotated frame, represented by numerical values, and explain its decision. Depending on the deployment platform, the model may also produce additional outputs, such as identifying actions that could lead to dangerous locations or recognizing objects in the scene.

\subsubsection{Controller}
Finally, the low-level, platform-specific controller executes the selected action on the platform.
In the case of the external camera scenario, the positions of the annotations chosen by the \gls{vlm} are used as vertices in a piecewise linear path.
The robot is then guided along this path using a PD controller measuring cross-track and heading error as demonstrated in \cite{robinson2023visual}.
In \gls{fpv} scenario, the controller is embedded into the environment and teleports the agent to the target location computed on the base of the input command.

\subsection{Adaptations to the FPV Setting}

We designed our framework with the AI2-THOR simulator \cite{kolve2022ai2thorinteractive3denvironment} as the platform, where the discrete robot actions comprise the robot's and camera's movements.
As shown in \cref{fig:1st_frame}, we enable the \gls{vlm} to interpret visual annotations that resemble a control overlay inspired by video-game interfaces. 
We display the numbers 1 to 7 on a semicircle at the bottom of the image, providing rotational control, where 4 is the neutral \texttt{MOVE\_FORWARD} and the others numbers represent various degrees of rotation.

Additionally, the model can select \texttt{LOOK\_UP} and \texttt{LOOK\_DOWN} commands, associated with the non-displayed numbers 8 and 9. Lastly, number 0 is associated with \texttt{DONE} command to end the episode.

Following a structured procedure, we build the Experiential Memory by manually navigating the robot in the AI2-THOR environment.
At each timestamp, the operator selects an appropriate action based on the current visual context -- i.e. a command number -- and provides a natural language explanation, simulating a ``think-aloud'' process.
For instance, upon observing a microwave on the left, the explanation would state: ``A microwave is visible on the left, so the system will steer slightly to the left.''
The annotated image, the selected command, and the corresponding explanation are inserted in the Experiential Memory.
This process aims to imitate human-like physical movements and capture the underlying thought processes that drive these actions.

\begin{figure}[t]
    \centering
    \includegraphics[width=0.85\linewidth]{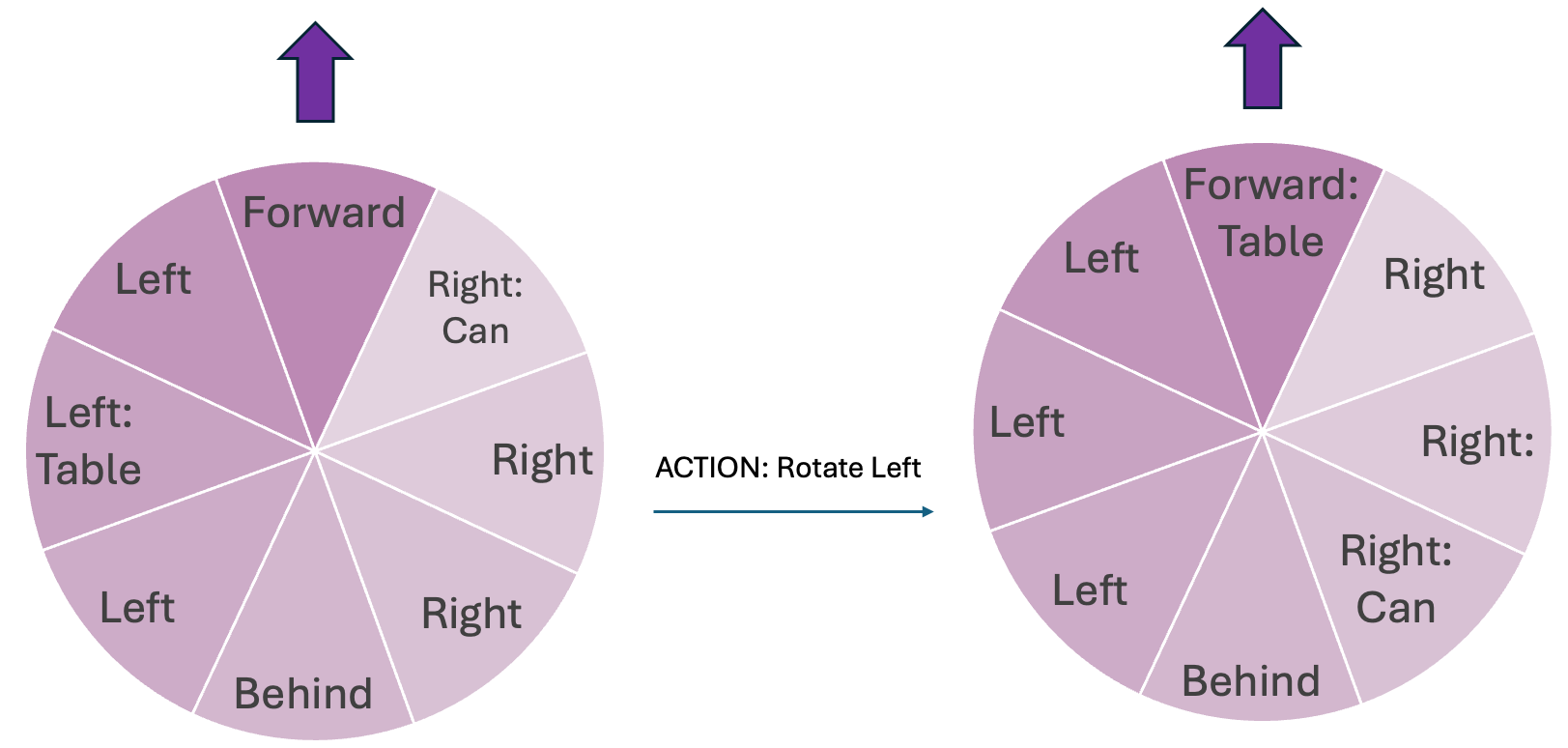}
    \caption{The figure depicts a scenario where the agent uses the compass.
    The compass keeps track of the scene content as the robot rotates, remembering insightful information about the room's layout.
    For instance, if the agent is looking for a \textit{chair}, it will likely rotate towards where it last saw a \textit{table}, although it is now out of sight.}
    \label{fig:fpv1}
    \vspace{-15pt}
\end{figure}

Finally, we request the \gls{vlm} a text description of the environment -- the list of objects in the frame -- and save it in the Episodic Memory.
From it, we create a ``circular compass'', as shown in \cref{fig:fpv1}, which rotates along with the agent's rotations and can inform the model's decisions since certain objects can be found near affine items or go out of the field of view due to motion.
We also include the last action, the current vertical view status and the previous command list: the \gls{vlm} at each timestamp is asked not only to provide the current action, but also the next future one, to robustly follow the navigation strategy in act (\cref{fig:1st_frame}).

\subsection{Adaptations to the TPV Setting}
\label{sec:method:3pv}

We follow the setup of \cite{robinson2023robot,robinson2023visual} in considering a visual servoing system \cite{liang2015adaptive,liang2020purely}, where a robot is controlled by the cloud via cameras installed in the infrastructure.
Here, security cameras capture raw frames of the environment, which are then annotated and passed, together with samples from the Experiential Memory, to the \gls{vlm} to predict the next few trajectory points for the robot to execute.
\Cref{fig:3rd_frame} visualises and summarises the framework's components.

We start the annotation process of the live frame by identifying the robot's position \cite{robinson2023robot}, through a YOLO model and place there the number 0 to make the task embodiment-agnostic.
During the system's initial setup, a segmentation mask of the scene's floor is taken through Segment Anything \cite{kirillov2023segment} and saved as a binary mask.
Importantly, this step is performed \textit{only once} during the framework's first setup, so it does not handle new objects or obstacles appearing in the scene.
Starting from the 0 position, we create concentric circles of numbers equally spaced and with increasing radius while using the mask, we filter out locations that are not traversable.
Finally, we ask the \gls{vlm} to choose a sequence of points from the robot to the end goal, which can be given as an object in the scene or a red circle -- clear of obstacles -- see \cref{fig:3rd_frame}.

An optional step is to crop the image to comprehend only the labels and the target to optimise the image size passed to the model, reducing costs and increasing focus on the important part of the picture.

\begin{figure*}
    \centering
    \begin{subfigure}[b]{0.26\textwidth}
        \includegraphics[width=\textwidth]{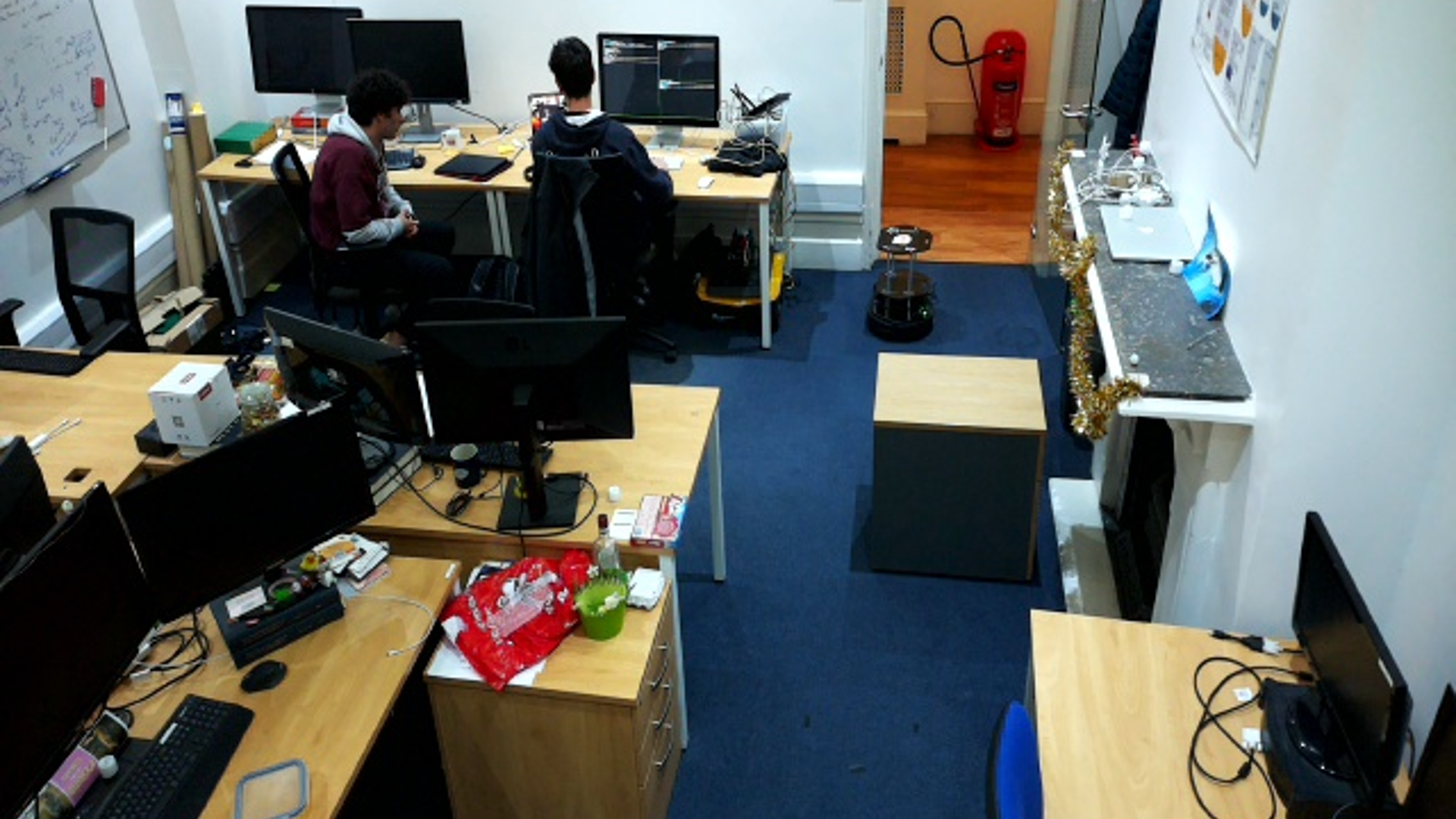}
    \end{subfigure}
    \begin{subfigure}[b]{0.26\textwidth}
        \includegraphics[width=\textwidth]{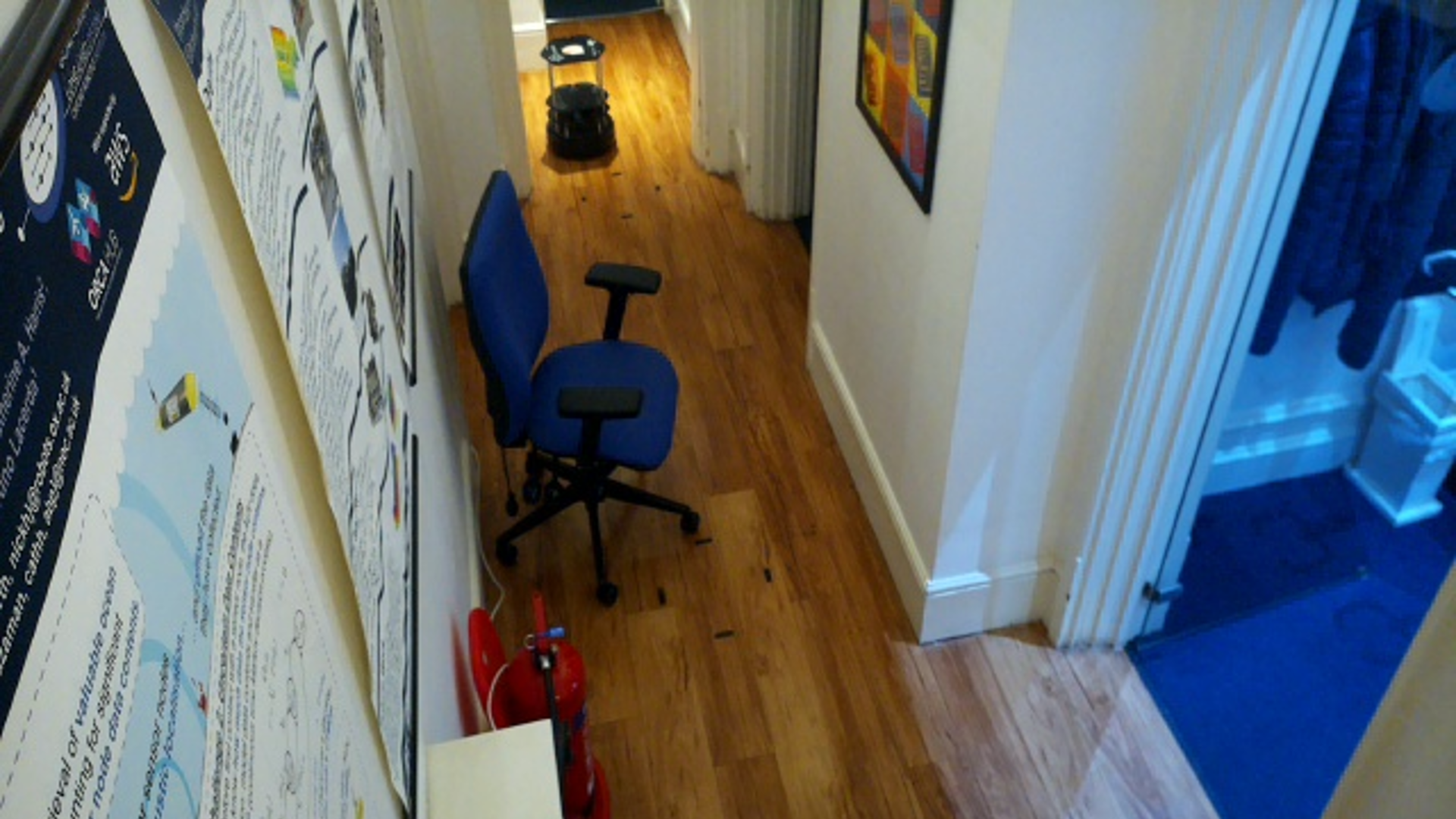}
    \end{subfigure}
    \begin{subfigure}[b]{0.26\textwidth}
        \includegraphics[width=\textwidth]{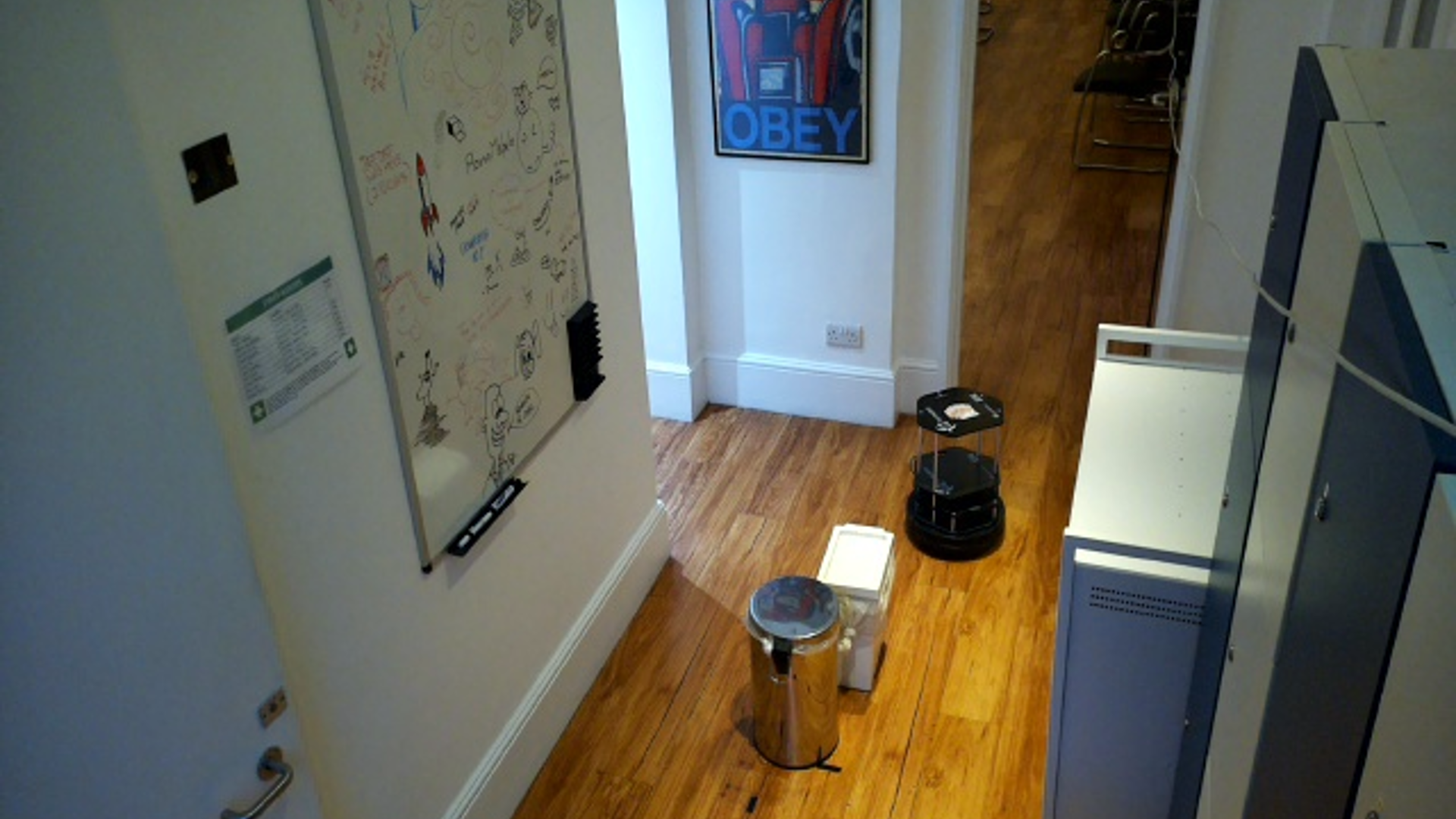}
    \end{subfigure}
    \caption{Examples rooms in the \gls{tpv} scenario. Random obstacles are placed to challenge the planner, e.g. the blue chair.}
    \label{rfidtag_testing}
    \vspace{-15pt}
\end{figure*}

After receiving the sequence of points, we map back the numbers to coordinates in image space and follow the path generated through a PID controller \cite{robinson2023visual}.
During inference, the sequence produced is at most 3 or 4 points long but once 1 or 2 points are correctly tracked, the new sequence substitutes the old one. 

\section{Experimental Setup}

We preliminary explored open-source and closed-source \glspl{vlm}; 
we empirically chose Gemini 1.5 Pro \cite{reid2024gemini} for its accessibility, performance and huge context-width (\texttt{gemini-1.5-pro-001}).
Hereinafter, with \gls{vlm} we implicitly refer to the latter.
The experimental setup differs slightly for \gls{fpv} and \gls{tpv}.
In the following, we will discuss both of them and describe the metrics, data collection and baselines we will use to evaluate our system.

\subsection{First Person View}
Among the several simulators proposed to facilitate the Embodied Navigation tasks \cite{savva2019habitatplatformembodiedai,kolve2022ai2thorinteractive3denvironment,xia2018gibsonenvrealworldperception}, we selected AI2-THOR's ObjectNav task, whose goal is to navigate towards a predefined target object.

We follow the evaluation procedure of \cite{wang2024goal}, using the same metrics and setup, with different object classes for training and testing\footnote{
\cite{wang2024goal} chose as training objects: \textit{HousePlant, Sink, TableTop, Knife, Fridge, Bowl, Cabinet, Cloth, KeyChain, WateringCan, Bed, Lamp, Book, Chair, LightSwitch, Candle, Painting, Watch, Cabinet, Toilet, SprayBottle} (for us, there is one episode for each one of them in the experiential memory). As test objects they chose: \textit{Toaster, Microwave, Television, LapTop, RemoteControl, CellPhone, Mirror, AlarmClock, Toiletpaper, SoapBottle}.}. We hence compare our approach to
\cite{wang2024goal}
\cite{yang2018visual}, \cite{wortsman2019learning}, \cite{du2021vtnet}.
Due to resource constraints, we evaluate our framework on 300 episodes per scene type, and limit maximum number of steps per episode to 25, thereby increasing the complexity of the task.
We manually navigate the robot through the environments, recording one episode per target object, resulting in a total of just 25 episodes.
This database is extremely limited purposefully to demonstrate that even few episodes can establish an effective framework, thereby challenging the generalization capabilities of trained models.

This constitutes a human-like sub-optimal ground truth, which composes the system's Experiential Memory and helps the model generate situation-grounded strategies.
We will compare our model against the same baseline models of \cite{wang2024goal}, to demonstrate how our framework compares with trained methods.
We will compare our model against the same baseline models
of [31], to demonstrate how our framework compares with
trained methods.
\subsection{Third Person View}

\begin{figure*}[h!]
    \centering
    \includegraphics[width=0.74\textwidth]{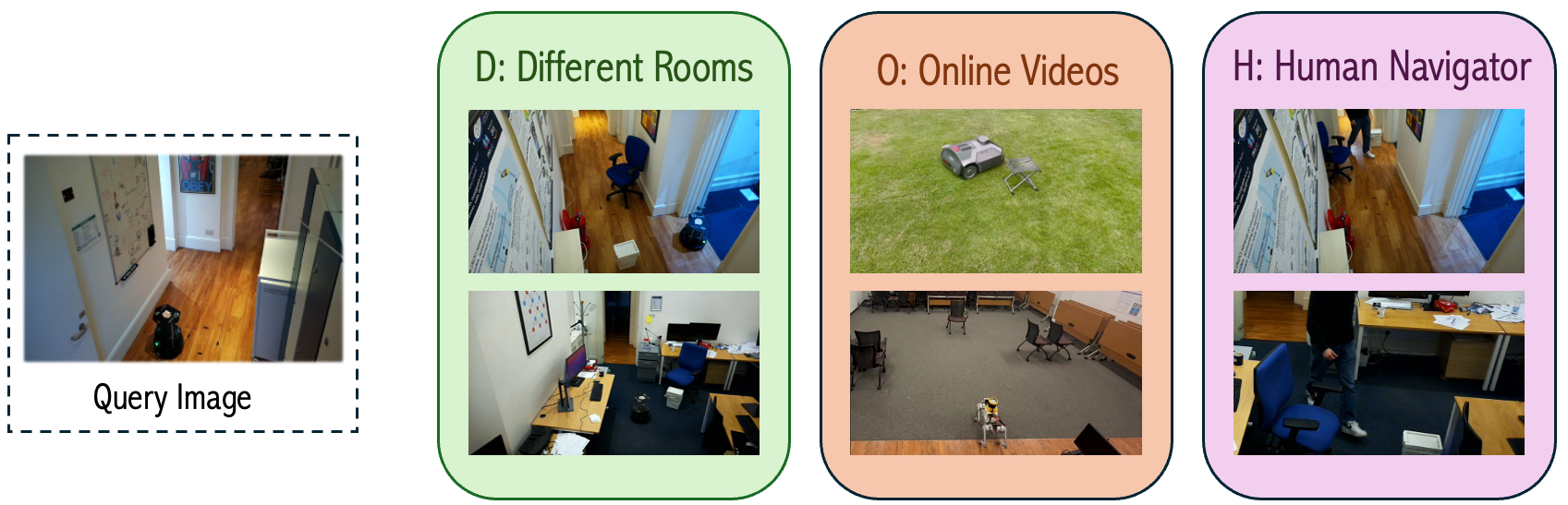}
    \caption{Experiential Memories for \gls{tpv}: \texttt{D} includes experiences from the same environment excluding the inference room, \texttt{O} from online videos and \texttt{H} from the same environment but with a human as navigator instead of a robot.}
    \label{fig:context}
    \vspace{-7pt}
\end{figure*}

We evaluate our approach on a custom dataset recorded in four rooms of the Oxford Robotics Institute premises -- see \cref{rfidtag_testing} -- where we collected expert trajectories by teleoperating a TurtleBot3 rover.
The trajectories recorded encompass a range of difficulty levels, from simple paths with minimal obstacles to more complex routes that include various dynamic and static obstacles, representing real-world navigation challenges.
We annotate -- see \cref{sec:method:3pv} -- the images and manually mark the labels overlapping with obstacles or non-traversable locations as \textit{dangerous}.
We compare the results of our model against a zero-shot approach on this dataset.
We test different Experiential Memories, composed of scenes from the same or different cameras, called scenarios \texttt{A} and \texttt{D}, respectively -- see \cref{fig:context}.
In addition, we will show results with trajectories performed by a human in the same scenes -- scenario \texttt{H} -- simulating the images usually captured by security cameras; this would demonstrate how a person move in an office room, avoiding obstacles such as chairs, boxes, etc.
Finally, we will also use short clips of robots from the web \cite{h1-video}, \cite{unitree-video} -- scenario \texttt{O} -- to validate how general and different from the target scenario the samples in the context can be, while still allowing the model to understand the task and mimic it successfully.

The main metric to evaluate our system is the Trajectory Score (TS), defined as:
\begin{equation}    
\mathrm{TS} = \sum_{i=0}^{N}S_i\frac{Pc_i}{\max(\text{len}(P_i), \text{len}(G_i))}
\hspace{0.5cm}  TS_i\in [0,1]
\end{equation}
where $Pc_i$ is the number of correct predicted points, $P_i$ is the predicted sequence and $G_i$ is the ground truth sequence, and $S_i$ indicates if the selected point is safe, at iteration $i$. The maximum value of this trajectory score is $N$, in our case 300, which is also the maximum TS value. The purpose of this metric is to measure to which extent the model is able to reproduce human-like navigation.
Moreover, we measure the number of dangerous points selected during the evaluation process.
\begin{equation}
    D = \sum_{i=0}^{N}D_i
        \hspace{0.5cm} D_i\in \{0,1\}
\end{equation}
where $D_i$ indicates if at episode $i$ a dangerous point has been selected or crossed, simulating a collision.

To make the evaluation goal-agnostic, we consider as end-goal a red circle, superimposed on the picture and report results averaging three consecutive runs.

\begin{table}[]
    \centering
    \renewcommand{\arraystretch}{1.2}
    \begin{tabular}{c|c|c|c|c}
        \textbf{Mode} & \textbf{CL} & \textbf{Scenario} &
        \textbf{TS (/300) $\uparrow$ } & \textbf{D (/300)}$ \downarrow$ \\
        \hline
        Zero-Shot & 0 & - & 147.82 & 76\\
        \hline
        \textbf{ICL (Ours)} & \textbf{10} & \textbf{A} & \textbf{270.70} & \textbf{2}\\
        \textbf{ICL (Ours)} & 10 & D & 247.24 & 8\\
        \textbf{ICL (Ours)} & 10 & H & 219.58 & 13\\
        \textbf{ICL (Ours)} & 10 & O & 235.83 & 16\\
    \end{tabular}
    \caption{\gls{tpv} results: comparison between the zero-shot and our framework on the same scenario but using different types of context sources (based on Scenario column).}
    \label{tab:tpv_res}
    \vspace{-15pt}
\end{table}

\section{Experimental Results}\label{sec:results}
\begin{table*}[htbp]
\centering
\begin{tabular}{llcccccc}
\multirow{2}{*}{\textbf{Eval. set}} & \multirow{2}{*}{\textbf{Method}} & \multicolumn{5}{c}{\textbf{SR $\|$ SPL \%}} \\
\cmidrule{3-7} 
& & \textbf{Kitchen} & \textbf{Living room} & \textbf{Bedroom} & \textbf{Bathroom} & \textbf{Average} & \\
\midrule
\multirow{5}{*}{\shortstack{Known scenes \\ \& Known objects}} 
& Random policy & 10.40 $\|$ \phantom{0}4.80 & 11.47 $\|$ \phantom{0}3.40 & 13.07 $\|$ \phantom{0}8.20 & 21.60 $\|$ 11.13 & 14.13 $\|$ \phantom{0}6.88\\
& Scene priors & 50.67 $\|$ 34.27 & 67.07 $\|$ 29.43 & 69.07 $\|$ 22.47 & 71.33 $\|$ 28.73 & 64.53 $\|$ 28.73 \\
& SAVN & 46.27 $\|$ 38.27 & 56.93 $\|$ 40.60 & 74.67 $\|$ 35.47 & 81.87 $\|$ 46.53 & 64.93 $\|$ 40.22 \\
& VTNET & 57.60 $\|$ 43.20 & 68.53 $\|$ 45.03 & 86.53 $\|$ 38.53 & 76.13 $\|$ 52.30 & 72.20 $\|$ 44.77  \\
& GVSN & \textbf{63.20} $\|$ \textbf{50.27} & \textbf{88.00} $\|$ \textbf{52.43} & \textbf{94.27} $\|$ \textbf{59.07} & \textbf{89.47} $\|$ \textbf{66.37} & \textbf{83.73} $\|$ \textbf{57.03}  \\
& Ours & 45.50 $\|$ 31.05 & 28.50 $\|$ 19.0  & 50.40 $\|$ 25.29 & 60.25 $\|$ 36.70 & 46.16 $\|$ 28.01 \\
\midrule
\multirow{5}{*}{\shortstack{Known scenes \\ \& Novel objects}} 
& Random policy & \phantom{0}2.93 $\|$ \phantom{0}0.66 & \phantom{0}5.60 $\|$ \phantom{0}0.93 & \phantom{0}8.80 $\|$ \phantom{0}2.53 & \phantom{0}8.13 $\|$ \phantom{0}2.03 & \phantom{0}6.37 $\|$ \phantom{0}1.54 \\
& Scene priors & 21.07 $\|$ 14.55 & 23.20 $\|$ \phantom{0}9.40 & 19.47 $\|$ 12.17 & 30.53 $\|$ 18.47 & 23.57 $\|$ 13.65 \\
& SAVN & 17.27 $\|$ \phantom{0}8.30 & 27.20 $\|$ \phantom{0}7.60 & 37.87 $\|$ 20.47 & 32.53 $\|$ 16.50 & 28.72 $\|$ 13.22 \\
& VTNET & 26.53 $\|$ 12.27 & 49.07 $\|$ 23.97 & 35.87 $\|$ 18.63 & 36.67 $\|$ 22.53 & 37.03 $\|$ 19.35 \\
& GVSN & \textbf{32.67} $\|$ \textbf{20.13} & \textbf{58.53} $\|$ \textbf{32.50} & \textbf{53.07} $\|$ \textbf{20.97} & 49.07 $\|$ 25.60 & \textbf{48.33} $\|$ \textbf{24.80} \\
& Ours & 30.00 $\|$ 19.01 & 23.00 $\|$ 18.70 & 44.00 $\|$ 32.44 & \textbf{54.00} $\|$ \textbf{28.51} & 37.75 $\|$ 24.66 \\
\midrule
\multirow{5}{*}{\shortstack{Novel scenes \\ \& Known objects}} 
& Random policy & \phantom{0}6.00 $\|$ \phantom{0}0.87 & \phantom{0}4.20 $\|$ \phantom{0}1.27 & \phantom{0}3.47 $\|$ \phantom{0}0.37 & \phantom{0}4.67 $\|$ \phantom{0}1.30 & \phantom{0}4.58 $\|$ \phantom{0}0.95 \\
& Scene priors & 11.60 $\|$ \phantom{0}6.23 & 13.87 $\|$ \phantom{0}8.27 & 17.20 $\|$ 10.83 & 15.07 $\|$ \phantom{0}8.40 & 14.43 $\|$ \phantom{0}8.43 \\
& SAVN & 26.80 $\|$ 10.70 & 31.33 $\|$ \phantom{0}6.00 & 43.87 $\|$ 15.63 & 21.73 $\|$ \phantom{0}8.33 & 30.93 $\|$ 10.17 \\
& VTNET & 35.73 $\|$ 12.30 & 40.93 $\|$ 13.93 & 57.87 $\|$ 17.83 & 47.60 $\|$ 10.73 & 45.53 $\|$ 13.70 \\
& GVSN & 44.13 $\|$ 18.50 & \textbf{48.00} $\|$ \textbf{27.53} & \textbf{68.67} $\|$ \textbf{19.50} & 60.53 $\|$ 26.07 & \textbf{55.33} $\|$ 22.90 \\
& Ours & \textbf{50.00} $\|$ \textbf{32.40} & 28.70 $\|$ 17.85 & 24.00 $\|$ 12.79 & \textbf{62.00} $\|$ \textbf{34.94} & 41.17 $\|$ \textbf{24.50} \\
\midrule
\multirow{5}{*}{\shortstack{Novel scenes \\ \& Novel objects}} 
& Random policy & \phantom{0}1.60 $\|$ \phantom{0}0.43 & \phantom{0}3.87 $\|$ \phantom{0}0.80 & \phantom{0}3.60 $\|$ \phantom{0}0.33 & \phantom{0}2.27 $\|$ \phantom{0}0.93 & \phantom{0}2.83 $\|$ \phantom{0}0.63 \\
& Scene priors & \phantom{0}2.93 $\|$ \phantom{0}1.13 & 10.80 $\|$ \phantom{0}3.13 & 23.07 $\|$ \phantom{0}7.60 & 15.87 $\|$ \phantom{0}6.37 & 13.17 $\|$ \phantom{0}4.56 \\
& SAVN & 17.33 $\|$ \phantom{0}5.97 & 13.60 $\|$ \phantom{0}4.50 & 25.47 $\|$ \phantom{0}5.50 & 12.27 $\|$ \phantom{0}4.37 & 17.17 $\|$ \phantom{0}5.08 \\
& VTNET & 26.67 $\|$ \phantom{0}7.03 & 19.47 $\|$ \phantom{0}9.03 & 16.93 $\|$ \phantom{0}7.40 & 16.93 $\|$ \phantom{0}7.40 & 22.43 $\|$ \phantom{0}6.79 \\
& GVSN & 34.40 $\|$ \phantom{0}7.30 & 17.87 $\|$ \phantom{0}8.10 & \textbf{32.40} $\|$ \phantom{0}9.60 & 30.00 $\|$ \phantom{0}6.57 & 28.67 $\|$ \phantom{0}7.89 \\
& Ours & \textbf{52.40} $\|$ \textbf{40.75} & \textbf{35.40} $\|$ \textbf{21.61} & 26.00 $\|$ \textbf{17.48} & \textbf{42.00} $\|$ \textbf{34.65} & \textbf{38.95} $\|$ \textbf{28.62} \\
\end{tabular}
\caption{Comparison of SR and SPL across different environments (kitchen, living room, bedroom, and bathroom). The table shows that S2P outperforms trained models in novel scenes, particularly in terms of average SR and SPL, indicating superior generalization across different environments and object configurations with minimal data needed.}
\label{tab:fvp_res}
\vspace{-15pt}
\end{table*}

Our \gls{icl}-based approach in the \gls{tpv} scenario achieved a maximum TS of 270.70 using context scenario \texttt{A}, which allowed unrestricted retrieval from the database.
This performance significantly outperformed the zero-shot approach -- by approximately 40\% -- demonstrating the robustness of our model when supported by a comprehensive context.

Moreover, the model exhibited a 24\% reduction in the selection of dangerous points, which would otherwise lead to potential collisions.
This reduction is a critical enhancement, as it directly correlates to safer navigation, an essential factor for real-world autonomous applications.

The model's capability to avoid hazardous locations was evident in scenario \texttt{A} but improved also across other contexts, including scenarios from online videos and human-driven trajectories.
This versatility underscores the model's adaptability and ability to generalize across varying sources of contextual information, demonstrating its robustness in understanding the task at hand.
These findings are significant because they illustrate that effective navigation can be achieved with minimal data collection, or even none, leveraging just online data, significantly reducing the costs and time associated with data gathering and model training, making it a practical solution for real-world deployment.

In the \gls{fpv} setup, S2P achieved an average Success Rate (SR) of 46.16\% in the scenario most favourable to trained models: \textit{known scenes and known objects}, where they unquestionably performed much better.
This was expected since our model retrieved from an extremely small database -- just 1 episode per object type -- while the best performing model \cite{wang2024goal} was trained on 8 millions of episodes. 
In more challenging scenarios, where generalization is required to handle novel scenes and objects, S2P consistently performed well, outperforming even the best one by approximately 10\% in Success Rate (SR) on average. Additionally, the Success weighted by Path Length (SPL) metric showed an improvement of around 20\% over the best-performing trained model in these scenarios. We hence reduce the number of samples needed by S2P to about 0.0005\% with respect to \cite{wang2024goal}.
These results validate the effectiveness of integrating \gls{icl} and \glspl{vlm} for navigation tasks.
Our approach achieves performance comparable to models trained extensively on specific tasks but with a handful of demonstrations.
By leveraging the pre-training of \glspl{vlm}, our method is also highly generalisable, capable of identifying different types of objects and adapting to various environmental settings.
This adaptability makes our framework suitable for various applications, from robotic navigation in unfamiliar environments to automated surveillance systems using external cameras.
\section{Conclusions and Future Works}
In conclusion, our experimental results underscore the potential of \gls{icl}-based frameworks combined with \glspl{vlm} for autonomous navigation. Our model significantly outperforms zero-shot baselines and adapts effectively to diverse contexts with minimal data and no specialized training. By intentionally operating under conditions of data scarcity, we demonstrated the robustness of the framework and anticipate further performance improvements as more episodes are incorporated.
This work establishes a strong foundation for scalable and flexible navigation systems, paving the way for more intelligent and adaptable autonomous technologies in real-world applications.
\bibliographystyle{IEEEtran}
\bibliography{main}{}

\begin{thebibliography}{10}
\providecommand{\url}[1]{#1}
\csname url@samestyle\endcsname
\providecommand{\newblock}{\relax}
\providecommand{\bibinfo}[2]{#2}
\providecommand{\BIBentrySTDinterwordspacing}{\spaceskip=0pt\relax}
\providecommand{\BIBentryALTinterwordstretchfactor}{4}
\providecommand{\BIBentryALTinterwordspacing}{\spaceskip=\fontdimen2\font plus
\BIBentryALTinterwordstretchfactor\fontdimen3\font minus \fontdimen4\font\relax}
\providecommand{\BIBforeignlanguage}[2]{{%
\expandafter\ifx\csname l@#1\endcsname\relax
\typeout{** WARNING: IEEEtran.bst: No hyphenation pattern has been}%
\typeout{** loaded for the language `#1'. Using the pattern for}%
\typeout{** the default language instead.}%
\else
\language=\csname l@#1\endcsname
\fi
#2}}
\providecommand{\BIBdecl}{\relax}
\BIBdecl

\bibitem{staroverov2020real}
A.~Staroverov, D.~A. Yudin, I.~Belkin, V.~Adeshkin, Y.~K. Solomentsev, and A.~I. Panov, ``Real-time object navigation with deep neural networks and hierarchical reinforcement learning,'' \emph{IEEE Access}, vol.~8, pp. 195\,608--195\,621, 2020.

\bibitem{visualrlzhou2021}
K.~Zhou, C.~Guo, and H.~Zhang, ``Visual navigation via reinforcement learning and relational reasoning,'' in \emph{2021 IEEE SmartWorld, Ubiquitous Intelligence \& Computing, Advanced \& Trusted Computing, Scalable Computing \& Communications, Internet of People and Smart City Innovation}, 2021, pp. 131--138.

\bibitem{shah2023gnm}
\BIBentryALTinterwordspacing
D.~Shah, A.~Sridhar, A.~Bhorkar, N.~Hirose, and S.~Levine, ``Gnm: A general navigation model to drive any robot,'' 2023. [Online]. Available: \url{https://arxiv.org/abs/2210.03370}
\BIBentrySTDinterwordspacing

\bibitem{zeng2024perceivereflectplandesigning}
\BIBentryALTinterwordspacing
Q.~Zeng, Q.~Yang, S.~Dong, H.~Du, L.~Zheng, F.~Xu, and Y.~Li, ``Perceive, reflect, and plan: Designing llm agent for goal-directed city navigation without instructions,'' 2024. [Online]. Available: \url{https://arxiv.org/abs/2408.04168}
\BIBentrySTDinterwordspacing

\bibitem{radford2019language}
A.~Radford, J.~Wu, R.~Child, D.~Luan, D.~Amodei, I.~Sutskever \emph{et~al.}, ``Language models are unsupervised multitask learners,'' \emph{OpenAI blog}, vol.~1, no.~8, p.~9, 2019.

\bibitem{williams2024masked}
\BIBentryALTinterwordspacing
D.~S.~W. Williams, M.~Gadd, P.~Newman, and D.~D. Martini, ``Masked gamma-ssl: Learning uncertainty estimation via masked image modeling,'' 2024. [Online]. Available: \url{https://arxiv.org/abs/2402.17622}
\BIBentrySTDinterwordspacing

\bibitem{williams2024mitigating}
D.~S.~W. Williams, D.~D. Martini, M.~Gadd, and P.~Newman, ``Mitigating distributional shift in semantic segmentation via uncertainty estimation from unlabeled data,'' \emph{IEEE Transactions on Robotics}, vol.~40, pp. 3146--3165, 2024.

\bibitem{hu2021lora}
E.~J. Hu, Y.~Shen, P.~Wallis, Z.~Allen-Zhu, Y.~Li, S.~Wang, L.~Wang, and W.~Chen, ``Lora: Low-rank adaptation of large language models,'' \emph{arXiv preprint arXiv:2106.09685}, 2021.

\bibitem{nasiriany2024pivot}
\BIBentryALTinterwordspacing
S.~Nasiriany, F.~Xia, W.~Yu, T.~Xiao, J.~Liang, I.~Dasgupta, A.~Xie, D.~Driess, A.~Wahid, Z.~Xu, Q.~Vuong, T.~Zhang, T.-W.~E. Lee, K.-H. Lee, P.~Xu, S.~Kirmani, Y.~Zhu, A.~Zeng, K.~Hausman, N.~Heess, C.~Finn, S.~Levine, and B.~Ichter, ``Pivot: Iterative visual prompting elicits actionable knowledge for vlms,'' 2024. [Online]. Available: \url{https://arxiv.org/abs/2402.07872}
\BIBentrySTDinterwordspacing

\bibitem{sathyamoorthy2024convoi}
A.~J. Sathyamoorthy, K.~Weerakoon, M.~Elnoor, A.~Zore, B.~Ichter, F.~Xia, J.~Tan, W.~Yu, and D.~Manocha, ``Convoi: Context-aware navigation using vision language models in outdoor and indoor environments,'' \emph{arXiv preprint arXiv:2403.15637}, 2024.

\bibitem{robinson2023robot}
L.~Robinson, M.~Gadd, P.~Newman, and D.~D. Martini, ``Robot-relay: Building-wide, calibration-less visual servoing with learned sensor handover networks,'' in \emph{International Symposium on Experimental Robotics}.\hskip 1em plus 0.5em minus 0.4em\relax Springer, 2023, pp. 129--140.

\bibitem{robinson2023visual}
L.~Robinson, D.~De~Martini, M.~Gadd, and P.~Newman, ``Visual servoing on wheels: Robust robot orientation estimation in remote viewpoint control,'' in \emph{2023 IEEE/RSJ International Conference on Intelligent Robots and Systems (IROS)}.\hskip 1em plus 0.5em minus 0.4em\relax IEEE, 2023, pp. 6364--6370.

\bibitem{zhong2024nerfoot}
D.~Zhong, L.~Robinson, and D.~De~Martini, ``Nerfoot: Robot-footprint estimation for image-based visual servoing,'' \emph{arXiv preprint arXiv:2408.01251}, 2024.

\bibitem{fukushima2022objectmemorytransfo}
\BIBentryALTinterwordspacing
R.~Fukushima, K.~Ota, A.~Kanezaki, Y.~Sasaki, and Y.~Yoshiyasu, ``Object memory transformer for object goal navigation,'' 2022. [Online]. Available: \url{https://arxiv.org/abs/2203.14708}
\BIBentrySTDinterwordspacing

\bibitem{shah2023vint}
\BIBentryALTinterwordspacing
D.~Shah, A.~Sridhar, N.~Dashora, K.~Stachowicz, K.~Black, N.~Hirose, and S.~Levine, ``Vint: A foundation model for visual navigation,'' 2023. [Online]. Available: \url{https://arxiv.org/abs/2306.14846}
\BIBentrySTDinterwordspacing

\bibitem{sridhar2023nomad}
\BIBentryALTinterwordspacing
A.~Sridhar, D.~Shah, C.~Glossop, and S.~Levine, ``Nomad: Goal masked diffusion policies for navigation and exploration,'' 2023. [Online]. Available: \url{https://arxiv.org/abs/2310.07896}
\BIBentrySTDinterwordspacing

\bibitem{zhou2024navgpt2unleashingnavigationalreasoning}
\BIBentryALTinterwordspacing
G.~Zhou, Y.~Hong, Z.~Wang, X.~E. Wang, and Q.~Wu, ``Navgpt-2: Unleashing navigational reasoning capability for large vision-language models,'' 2024. [Online]. Available: \url{https://arxiv.org/abs/2407.12366}
\BIBentrySTDinterwordspacing

\bibitem{duan2024manipulateanything}
\BIBentryALTinterwordspacing
J.~Duan, W.~Yuan, W.~Pumacay, Y.~R. Wang, K.~Ehsani, D.~Fox, and R.~Krishna, ``Manipulate-anything: Automating real-world robots using vision-language models,'' 2024. [Online]. Available: \url{https://arxiv.org/abs/2406.18915}
\BIBentrySTDinterwordspacing

\bibitem{pmlr-v205-shah23b}
\BIBentryALTinterwordspacing
D.~Shah, B.~Osi\'nski, b.~ichter, and S.~Levine, ``Lm-nav: Robotic navigation with large pre-trained models of language, vision, and action,'' in \emph{Proceedings of The 6th Conference on Robot Learning}, ser. Proceedings of Machine Learning Research, K.~Liu, D.~Kulic, and J.~Ichnowski, Eds., vol. 205.\hskip 1em plus 0.5em minus 0.4em\relax PMLR, 14--18 Dec 2023, pp. 492--504. [Online]. Available: \url{https://proceedings.mlr.press/v205/shah23b.html}
\BIBentrySTDinterwordspacing

\bibitem{chen2024mapgpt}
\BIBentryALTinterwordspacing
J.~Chen, B.~Lin, R.~Xu, Z.~Chai, X.~Liang, and K.-Y.~K. Wong, ``Mapgpt: Map-guided prompting with adaptive path planning for vision-and-language navigation,'' 2024. [Online]. Available: \url{https://arxiv.org/abs/2401.07314}
\BIBentrySTDinterwordspacing

\bibitem{dipalo2024kat}
\BIBentryALTinterwordspacing
N.~D. Palo and E.~Johns, ``Keypoint action tokens enable in-context imitation learning in robotics,'' 2024. [Online]. Available: \url{https://arxiv.org/abs/2403.19578}
\BIBentrySTDinterwordspacing

\bibitem{liu2021swintransformerhierarchicalvision}
\BIBentryALTinterwordspacing
Z.~Liu, Y.~Lin, Y.~Cao, H.~Hu, Y.~Wei, Z.~Zhang, S.~Lin, and B.~Guo, ``Swin transformer: Hierarchical vision transformer using shifted windows,'' 2021. [Online]. Available: \url{https://arxiv.org/abs/2103.14030}
\BIBentrySTDinterwordspacing

\bibitem{mmr}
J.~Carbonell and J.~Stewart, ``The use of mmr, diversity-based reranking for reordering documents and producing summaries,'' \emph{SIGIR Forum (ACM Special Interest Group on Information Retrieval)}, 06 1999.

\bibitem{kolve2022ai2thorinteractive3denvironment}
\BIBentryALTinterwordspacing
E.~Kolve, R.~Mottaghi, W.~Han, E.~VanderBilt, L.~Weihs, A.~Herrasti, M.~Deitke, K.~Ehsani, D.~Gordon, Y.~Zhu, A.~Kembhavi, A.~Gupta, and A.~Farhadi, ``Ai2-thor: An interactive 3d environment for visual ai,'' 2022. [Online]. Available: \url{https://arxiv.org/abs/1712.05474}
\BIBentrySTDinterwordspacing

\bibitem{liang2015adaptive}
X.~Liang, H.~Wang, W.~Chen, D.~Guo, and T.~Liu, ``Adaptive {Image}-{Based} {Trajectory} {Tracking} {Control} of {Wheeled} {Mobile} {Robots} {With} an {Uncalibrated} {Fixed} {Camera},'' \emph{IEEE Transactions on Control Systems Technology}, vol.~23, no.~6, pp. 2266--2282, Nov. 2015, conference Name: IEEE Transactions on Control Systems Technology.

\bibitem{liang2020purely}
X.~Liang, H.~Wang, Y.-H. Liu, Z.~Liu, B.~You, Z.~Jing, and W.~Chen, ``Purely {Image}-{Based} {Pose} {Stabilization} of {Nonholonomic} {Mobile} {Robots} {With} a {Truly} {Uncalibrated} {Overhead} {Camera},'' \emph{IEEE Transactions on Robotics}, vol.~36, no.~3, pp. 724--742, Jun. 2020, conference Name: IEEE Transactions on Robotics.

\bibitem{kirillov2023segment}
\BIBentryALTinterwordspacing
A.~Kirillov, E.~Mintun, N.~Ravi, H.~Mao, C.~Rolland, L.~Gustafson, T.~Xiao, S.~Whitehead, A.~C. Berg, W.-Y. Lo, P.~Dollár, and R.~Girshick, ``Segment anything,'' 2023. [Online]. Available: \url{https://arxiv.org/abs/2304.02643}
\BIBentrySTDinterwordspacing

\bibitem{reid2024gemini}
M.~Reid, N.~Savinov, D.~Teplyashin, D.~Lepikhin, T.~Lillicrap, J.-b. Alayrac, R.~Soricut, A.~Lazaridou, O.~Firat, J.~Schrittwieser \emph{et~al.}, ``Gemini 1.5: Unlocking multimodal understanding across millions of tokens of context,'' \emph{arXiv preprint arXiv:2403.05530}, 2024.

\bibitem{savva2019habitatplatformembodiedai}
\BIBentryALTinterwordspacing
M.~Savva, A.~Kadian, O.~Maksymets, Y.~Zhao, E.~Wijmans, B.~Jain, J.~Straub, J.~Liu, V.~Koltun, J.~Malik, D.~Parikh, and D.~Batra, ``Habitat: A platform for embodied ai research,'' 2019. [Online]. Available: \url{https://arxiv.org/abs/1904.01201}
\BIBentrySTDinterwordspacing

\bibitem{xia2018gibsonenvrealworldperception}
\BIBentryALTinterwordspacing
F.~Xia, A.~Zamir, Z.-Y. He, A.~Sax, J.~Malik, and S.~Savarese, ``Gibson env: Real-world perception for embodied agents,'' 2018. [Online]. Available: \url{https://arxiv.org/abs/1808.10654}
\BIBentrySTDinterwordspacing

\bibitem{wang2024goal}
Z.~Wang and G.~Tian, ``Goal-oriented visual semantic navigation using semantic knowledge graph and transformer,'' \emph{IEEE Transactions on Automation Science and Engineering}, 2024.

\bibitem{yang2018visual}
W.~Yang, X.~Wang, A.~Farhadi, A.~Gupta, and R.~Mottaghi, ``Visual semantic navigation using scene priors,'' \emph{arXiv preprint arXiv:1810.06543}, 2018.

\bibitem{wortsman2019learning}
M.~Wortsman, K.~Ehsani, M.~Rastegari, A.~Farhadi, and R.~Mottaghi, ``Learning to learn how to learn: Self-adaptive visual navigation using meta-learning,'' in \emph{Proceedings of the IEEE/CVF conference on computer vision and pattern recognition}, 2019, pp. 6750--6759.

\bibitem{du2021vtnet}
H.~Du, X.~Yu, and L.~Zheng, ``Vtnet: Visual transformer network for object goal navigation,'' \emph{arXiv preprint arXiv:2105.09447}, 2021.

\bibitem{h1-video}
\BIBentryALTinterwordspacing
H.~Robotics. H1 pro obstacle avoidance demo. Youtube. [Online]. Available: \url{https://www.youtube.com/watch?v=FgftdWrSYzM}
\BIBentrySTDinterwordspacing

\bibitem{unitree-video}
\BIBentryALTinterwordspacing
R.~Roy. {Unitree Go1 Obstacle Avoidance using Nav2 and 3D SLAM (RTAB MAP)}. [Online]. Available: \url{https://www.youtube.com/watch?v=hL4MTG0u1K0}
\BIBentrySTDinterwordspacing

\end{thebibliography}
\end{document}